\documentclass[runningheads]{llncs}
\usepackage[T1]{fontenc}
\usepackage{graphicx}
\usepackage{amssymb}
\usepackage{amsmath}
\usepackage{booktabs}
\usepackage[linesnumbered,ruled,vlined]{algorithm2e}
\usepackage{color}
\usepackage{booktabs}
\usepackage{multirow}
\usepackage{amssymb}
\begin{document}

\newcommand{\name}{\emph{CLIP-AGIQA}}
\title{CLIP-AGIQA: Boosting the Performance of AI-Generated Image Quality Assessment with CLIP}
\titlerunning{CLIP-AGIQA}
\author{Zhenchen Tang$^{\tiny \dagger}$ \and
Zichuan Wang$^{\tiny \dagger}$ \and
Bo Peng$^{\tiny *}$ \and Jing Dong$^{\tiny *}$}
\authorrunning{Z. Tang et al.}
\institute{New Laboratory of Pattern Recognition, Institute of Automation, Chinese Academy of Sciences \\ 
\email{\{tangzhenchen2024,wangzichuan2024\}@ia.ac.cn, \{bo.peng,jdong\}@nlpr.ia.ac.cn}\\}

\maketitle 
\renewcommand{\thefootnote}{\fnsymbol{footnote}} %将脚注符号设置为fnsymbol类型，即特殊符号表示
\footnotetext[4]{These authors contributed equally.}
\footnotetext[1]{Corresponding authors.} 
\begin{abstract}
With the rapid development of generative technologies, AI-Generated Images (AIGIs) have been widely applied in various aspects of daily life. However, due to the immaturity of the technology, the quality of the generated images varies, so it is important to develop quality assessment techniques for the generated images. Although some models have been proposed to assess the quality of generated images, they are inadequate when faced with the ever-increasing and diverse categories of generated images. Consequently, the development of more advanced and effective models for evaluating the quality of generated images is urgently needed. Recent research has explored the significant potential of the visual language model CLIP in image quality assessment, finding that it performs well in evaluating the quality of natural images. However, its application to generated images has not been thoroughly investigated. In this paper, we build on this idea and further explore the potential of CLIP in evaluating the quality of generated images. We design \name, a CLIP-based regression model for quality assessment of generated images, leveraging rich visual and textual knowledge encapsulated in CLIP. Particularly, we implement multi-category learnable prompts to fully utilize the textual knowledge in CLIP for quality assessment. Extensive experiments on several generated image quality assessment benchmarks, including AGIQA-3K and AIGCIQA2023, demonstrate that \name\ outperforms existing IQA models, achieving excellent results in evaluating the quality of generated images.

\keywords{AI-Generated Images  \and CLIP \and Perceptual Quality.}
\end{abstract}
\section{Introduction}
With the rapid development of generative technologies, Artificial Intelligence Generated Images (AIGIs) have become increasingly ubiquitous in modern society. From avatar generation on social media to visual effects production in movies and television, and even content creation in virtual and augmented 
\begin{figure}
\includegraphics[width=\textwidth]{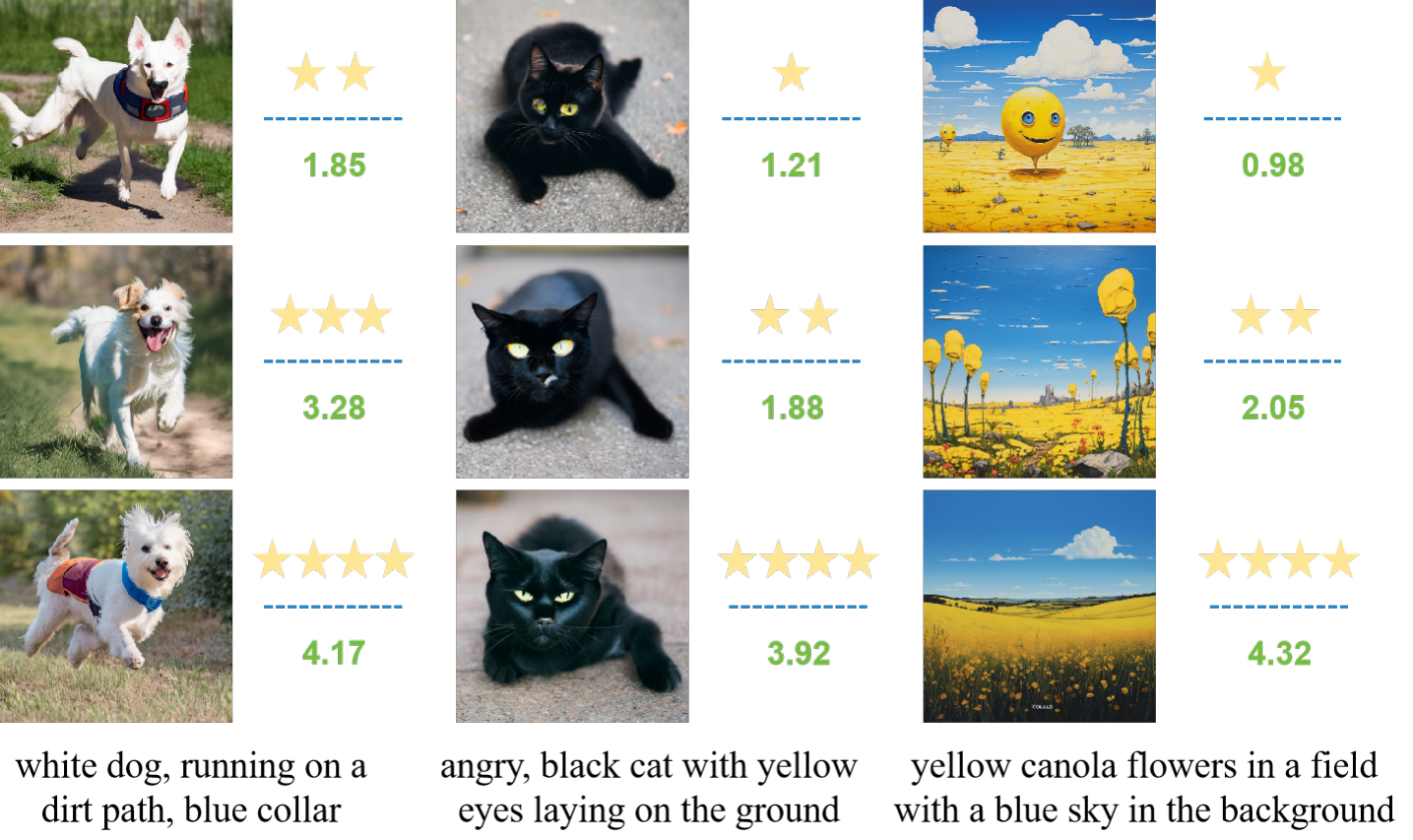}
\caption{Performance of \name. The star icons represent human ratings, and the green scores below the dashed line represent the scores predicted by our model.} \label{fig1}
\end{figure}
reality, generative technologies has become an integral part of our daily experiences. However, alongside these technological advancements, assessing the quality of generated images has become an emerging issue. Due to the immaturity of the technology, the quality of generated images is uneven, which can lead to unsatisfactory user experiences in some applications~\cite{li2023agiqa}. Therefore, developing techniques to effectively evaluate the quality of generated images is particularly important.

Quality assessment of generated images involves evaluating various dimensions through subjective and objective methods, such as the perceptual quality and the content accuracy with respect to input prompts. Recent efforts have focused on creating comprehensive databases for subjective quality assessment based on human perception and developing approaches to enhance evaluation performance~\cite{li2023agiqa,xu2024imagereward,kirstain2023pick}. Despite these advancements, existing methods struggle to keep pace with the increasing diversity of generated images. For instance, in the field of text-to-image (T2I) generative models alone, there have been at least 20 representative T2I AGI models up to 2023, as indicated by recent statistics~\cite{frolov2021adversarial,zhang2023text}. Therefore, more research is needed to meet the quality assessment demands in this field.

Recent research has begun to explore CLIP's~\cite{radford2021learning} (Contrastive Language-Image Pre-training) potential in image quality assessment, revealing its effectiveness in evaluating natural images~\cite{wang2023exploring}. CLIP demonstrates strong performance across various visual and multimodal tasks due to its extensive pre-training on language-image data. However, since CLIP is pre-trained on natural images, it may have problems to model the quality distribution of generated images effectively, leaving a gap in this area. To address this, we propose \name, a CLIP-based regression model that leverages CLIP's comprehensive visual and textual knowledge to evaluate the quality of generated images. First, we design various prompts representing different quality levels to input into CLIP's text encoder, mitigating semantic ambiguities. Second, by introducing a learnable prompts strategy and utilizing multiple quality-related auxiliary prompts, we make full use of CLIP's textual knowledge. Last, our regression network then maps CLIP features to quality scores, effectively adapting CLIP's capabilities to the task of generated image quality assessment, thereby enhancing the model's performance. 
The specific performance of our \name\ can be seen in Fig.\ref{fig1}.

In summary, our primary contributions include:

\begin{itemize}
\item We propose \name, adapting the CLIP model to the task of evaluating generated image quality;
\item We introduce a learnable prompts strategy and design multiple prompts of varying quality levels to fully utilize CLIP's textual knowledge for assisting in evaluating generated image quality;
\item We conduct experiments on several benchmarks for generated image quality assessment such as AGIQA-3K and AIGCIQA2023, achieving state-of-the-art performance.
\end{itemize}

\section{Related Work}
\subsection{Image Quality Assessment}
Traditional image quality assessment aims to evaluate the quality of natural images, including aspects like noise, blur, compression artifacts, etc~\cite{gu2020giqa}. It is categorized into three types: full-reference, reduced-reference, and no-reference. Full-reference methods compare the original and test images, commonly using metrics like PSNR and SSIM~\cite{wang2004image}. Reduced-reference methods utilize partial information from a reference image, such as RRED~\cite{soundararajan2011rred} and OSVP~\cite{wu2016orientation}. No-reference methods directly assess image quality using machine learning and deep learning techniques, such as BRISQUE~\cite{mittal2012no}, IQA-CNN~\cite{kang2014convolutional} and RankIQA~\cite{liu2017rankiqa}.

In recent years, with the development of generative technologies, assessing the quality of generated images has become increasingly important. Due to potential abnormal distortions or unrealistic structures in generated images, evaluation focuses on visual perception, including authenticity, naturalness, and coherence. Common metrics include Inception Score (IS) for assessing image quality and diversity based on classification results and KL divergence~\cite{salimans2016improved}, Fréchet Inception Distance (FID) for evaluating visual quality by comparing feature distributions of real and generated images~\cite{heusel2017gans}, and CLIP Score, which assesses image quality based on similarity between generated images and textual descriptions~\cite{hessel2021clipscore}.

Recently, datasets like AGIQA-3K~\cite{li2023agiqa} and PKU-I2IQA~\cite{yuan2023pku} have been proposed to facilitate benchmark experiments for IQA models, focusing on the quality assessment of generated images. AGIQA-3K provides a comprehensive and diverse subjective quality database covering various generated images from GAN, autoregressive, and diffusion models. PKU-I2IQA, the first image-to-image AIGC quality assessment database based on human perception, also conducts benchmark experiments on different IQA models. Additionally, models such as ImageReward~\cite{xu2024imagereward} and HPS~\cite{wu2023human} construct datasets for generated images from the perspective of human preferences and proposed corresponding evaluation models, providing a benchmark for quality assessment in terms of human preferences for generated images. Despite these advancements, there remains a scarcity of specialized models for assessing the quality of generated images, necessitating further research to advance this field.
\subsection{CLIP-Based Methods}
CLIP~\cite{radford2021learning} is a large-scale vision-language pretrained model that leverages contrastive learning to achieve cross-modal knowledge understanding. It has demonstrated strong transfer capabilities across various visual tasks such as semantic segmentation (LSeg~\cite{li2022language}), object detection (ViLD~\cite{gu2021open}), and image generation (CLIPasso~\cite{vinker2022clipasso}). \par
CLIP-IQA~\cite{wang2023exploring} is the first work to explore CLIP in image quality assessment tasks, demonstrating that CLIP can be effectively extended to image quality evaluation. Due to the significant impact of linguistic ambiguity in quality assessment tasks~\cite{khurana2023natural}, phrases such as "a rich image" can be particularly problematic. This phrase could either refer to an image with rich content or an image associated with wealth. CLIP-IQA design an antonym prompt strategy to leverage CLIP's prior knowledge. However, due to the limited variety of prompts, this approach can result in inaccurate quality predictions. Moreover, this work only explored the performance of CLIP in natural image quality assessment tasks and did not address generated images. Building on this idea, we further investigate the performance of CLIP in evaluating the quality of generated images and propose a CLIP-based quality assessment regression model. By simultaneously fine-tuning our designed multi-class learnable prompts and the regression network added after CLIP, we achieve superior performance in assessing the quality of generated images.\par
Notably, recent methods~\cite{hou2023towards,ke2023vila,zhang2023blind} also explore CLIP for IQA, with many focusing on aesthetic evaluation. These methods stand out for their pioneering efforts in multi-modality integration for low-level vision and their impressive zero-shot performance. However, since CLIP is pre-trained on natural image-text pairs, directly using CLIP in a zero-shot manner to evaluate the quality of generated images, as done in the aforementioned methods, does not yield optimal results. Therefore, we train a CLIP-based model using generated images to better model the quality distribution of generated images.

\section{Methodology}
In this section, we first formalize the paradigm of a typical IQA model. Then, we provide a detailed description of the various designs we implement to adapt CLIP to the task of generative image quality assessment in \name.

\subsection{Preliminary on IQA Models}
Given an image $I$, a typical IQA model uses a visual encoder $V(\cdot)$ to extract visual features, followed by a regression model $R(\cdot)$ to predict the quality score. This process can be represented as follows:\par
\begin{equation}
S = R(V(I))
\end{equation}
In CLIP-IQA~\cite{wang2023exploring}, only the visual encoder \( V(\cdot) \) is used to extract visual features, and then an antonym prompt strategy is employed to compute the cosine similarity with the visual features to predict the quality score. Specifically, CLIP-IQA adopts antonym prompts (e.g., "Good photo." and "Bad photo.") as a pair for each prediction. Let \( x \) represent the features from the image, and \( t_1 \) and \( t_2 \) be the features from the two prompts with opposite meanings. The cosine similarity is computed as follows~\cite{wang2023exploring}:

\begin{equation}
s_i = \frac{x \cdot t_i}{||x|| \cdot ||t_i||}, \quad i \in \{1, 2\},
\end{equation}

and Softmax is used to compute the final score \( \bar{s} \in [0, 1] \):

\begin{equation}
\bar{s} = \frac{e^{s_1}}{e^{s_1} + e^{s_2}}.
\end{equation}
When a pair of adjectives is used, the ambiguity of one prompt is reduced by its antonym, casting the task as a binary classification where the final score is regarded as a relative similarity~\cite{wang2023exploring}.
Although this method effectively leverages the prior knowledge of CLIP, the predicted quality score is solely dependent on the contrastive similarity, which is not accurate. Therefore, in our design, we improve the network by using a regression model $R(\cdot)$ to predict the quality score, enhancing the precision of the prediction and better adapting CLIP to the quality assessment task after further reducing ambiguity with more fine-grained quality-related adjectives.

\subsection{Overview of \name}\label{sub1}
\begin{figure}
\includegraphics[width=\textwidth]{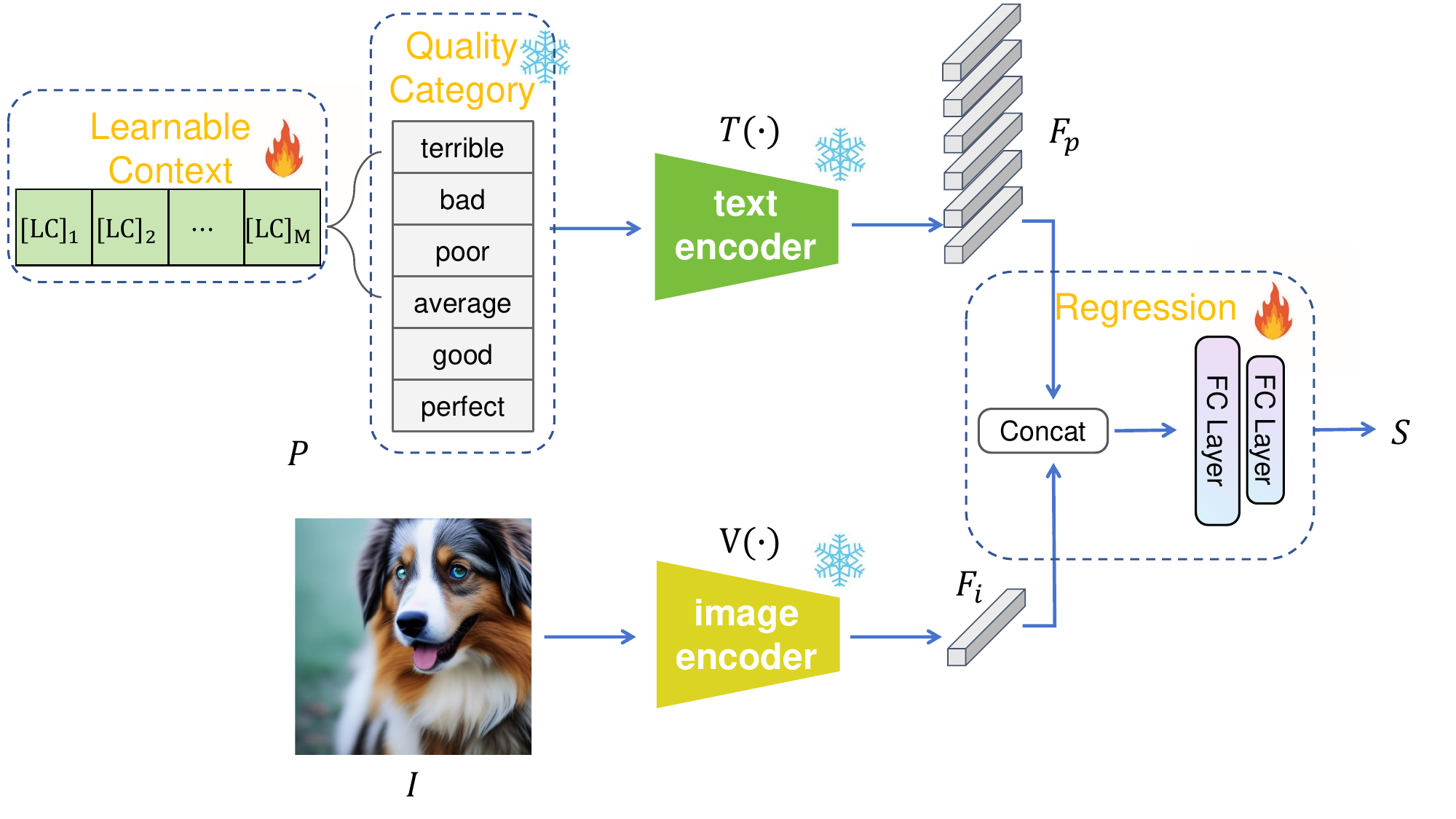}
\caption{Overall Architecture of \name.} \label{fig2}
\end{figure}

The overall framework of our method is shown in Fig.\ref{fig2}. \name\ consists of four components: learnable context, quality category, image quality regression, and the text encoder and image encoder in CLIP. In addition to the regression design, to better utilize the prior knowledge of the CLIP model, we incorporate learnable context for fine-tuning, inspired by the CoOp approach~\cite{zhou2022learning}. We also introduce additional quality category to address the ambiguity issues mentioned in CLIP-IQA. These two types of text-related information together form supplementary textual information to assist CLIP in adapting to the task of generative image quality assessment.

\subsubsection{Learnable Context.}
Since prompt engineering is a significant challenge in the application of CLIP, and the design of prompts can greatly impact performance, even with extensive manual tuning, the resulting prompts are by no means guaranteed to be optimal for downstream tasks~\cite{zhou2022learning}. Therefore, we abandon traditional subjective prompt settings in favor of a learnable prompt strategy. CLIP is sensitive to the choice of prompts, so we need to design a suitable set to leverage its prior knowledge. Similar to CoOp~\cite{zhou2022learning}, we avoid manual prompt adjustments by modeling the context words using continuous vectors, which are end-to-end learned from the data, while freezing a large number of CLIP's pre-trained parameters. Specifically, as shown in Fig.\ref{fig2}, we use learnable context. We employ a unified context version from CoOp, where all prompts share the same context. The prompt design for the text encoder \( T(\cdot) \) is as follows:

\begin{equation}
P = [LC]_1[LC]_2\ldots[LC]_M[QC] \label{eq1}
\end{equation}

Each \( [LC]_m \) (\( m \in \{1, \ldots, M\} \)) is the learnable context, represented as a vector with the same dimensionality as the word embeddings (i.e., 512 for CLIP). Here, \( M \) is a hyperparameter specifying the number of context tokens.

\subsubsection{Text Encoder And Image Encoder.}
We utilize the text encoder $T(\cdot)$ and image encoder $V(\cdot)$ from CLIP. The text encoder is based on a Transformer architecture ~\cite{vaswani2017attention} and is responsible for generating text representations from natural language. In contrast, the image encoder is designed to map high-dimensional images into a low-dimensional embedding space. This encoder's architecture can resemble a CNN like ResNet-50 ~\cite{he2016deep} or a Vision Transformer (ViT) ~\cite{dosovitskiy2020image}. In our setup, we employ these encoders separately to process our input textual information \(P\) and image information \(I\), generating intermediate features used to predict quality score.

\subsubsection{Quality Category.}
Due to the inherent language ambiguity in quality assessment tasks, utilizing CLIP as a versatile prior for visual perceptual evaluation is not straightforward. Similar to the antonym design in CLIP-IQA, we employ a series of quality-related auxiliary categories in Equation (\ref{eq1}) \([QC]\) to enhance the expression of the quality assessment task by describing the goodness of quality in a finer granularity. When using a set of quality-related adjective categories, they align with the correct category akin to the antonym prompts in CLIP-IQA, thereby reducing ambiguity. This transforms the task into multi-class classification, where the final score can be regarded as relative similarity, calculated through regression rather than using softmax as in CLIP-IQA. Specifically, we utilize six adjectives—terrible, bad, poor, average, good, and perfect—as quality category words to reduce ambiguity, thus better leveraging CLIP's priors. In addition, we also explore in the Section \ref{sub2} the impact of the number and types of different words on its effectiveness. This design, together with the setting of the first learnable context, constitutes additional textual information to assist CLIP in transferring to the task of generated image quality assessment.

\subsubsection{Image Quality Regression.}
To better fit the CLIP features to the data distribution for the task of evaluating the quality scores of generated images, we follow the paradigm of general quality assessment tasks by using the regression model \( R(\cdot) \) to predict quality scores. We concatenate the image features \( F_i = V(I) \in \mathbb{R}^{1\times N} \) and the textual features \( F_p = T(P) \in \mathbb{R}^{6\times N} \) as the input features \( F \). 
\begin{equation}
F = concat(F_i, F_p)
\end{equation}
We then process the concatenated features \( F \) through two fully connected (FC) layers. Here, the parameters of the FC layers are also learnable. The projection sizes are from\ 7 * 512\ to 512 and from 512 to 1, respectively. Finally, we obtain the predicted quality score \( S \), expressed as follows:

\begin{equation}
S = R(F)
\end{equation}
Throughout the entire learning process, we employ the Mean Squared Error (MSE) as the loss function, with the specific formula shown below:
\begin{equation}
L = \frac{1}{N} \sum_{i=1}^{n}(S-y)^2
\end{equation}
where $S$ represents the predicted quality score, and 
$y$ represents the ground truth of the quality score.

\section{Experiments}
\subsection{Experimental Settings}
\subsubsection{Datasets.}
To validate the effectiveness of our method, we conduct evaluations on two quality assessment benchmarks for generated images: AGIQA-3K~\cite{li2023agiqa} ans AIGCIQA2023~\cite{wang2023aigciqa2023}. AGIQA-3K is a database containing 2,982 AI-generated images produced by six different models, including GAN-based, auto-regression-based, and diffusion-based models and subjective experiments are organized to obtain MOS (Mean Opinion Score) labels in terms of perceptual quality, which range from 0 to 5. AIGCIQA2023 collects over 2000 images using 100 prompts and six state-of-the-art text-to-image generation models, and quality and authenticity ratings are obtained by subjective experiments, which are ultimately scaled to a range of 0-100.
\subsubsection{Evaluation Metrics.}
We use three common metrics in image quality assessment: PLCC, SRCC, and KRCC. PLCC (Pearson Linear Correlation Coefficient) measures the linear relationship between the predicted quality scores and the subjective scores. SRCC (Spearman Rank Correlation Coefficient) measures the consistency in the ranking order between the predicted quality scores and the subjective scores. KRCC (Kendall Rank Correlation Coefficient) measures the consistency in pairwise comparisons between the predicted quality scores and the subjective scores. All three metrics range from [-1, 1], with values closer to 1 indicating higher correlation.
\subsubsection{Training Details.}
The proposed \name\ is implemented in PyTorch and trained on 1 NVIDIA A100 GPU. ViT-B/16~\cite{dosovitskiy2020image} is used as the image encoder's backbone, and SGD is applied to optimize the network with an initial learning rate of 0.002. The training process was conducted over 100 epochs with a batch size of 32 and a learnable context length of 16. For learning rate scheduling, we employed a cosine annealing strategy, allowing the learning rate to decrease gradually throughout the training. Additionally, we implemented a warm-up phase during the first epoch, where the learning rate was held constant at \(1 \times 10^{-5}\). 

\subsection{Experiment on Different Datasets}
We focus on exploring the potential of \name\ in overall quality perception assessment. We conduct experiments on two widely used AGIQA benchmarks: AGIQA-3K~\cite{li2023agiqa} ans AIGCIQA2023~\cite{wang2023aigciqa2023}. We also compare \name\ with different IQA methods, including handcrafted-based methods such as CEIQ~\cite{yan2019no}, NIQE~\cite{mittal2012making} and BRISQUE~\cite{mittal2012no}, and several learning-based methods like DBCNN~\cite{zhang2018blind}, CLIP-IQA~\cite{wang2023exploring} and CNNIQA~\cite{kang2014convolutional}.

Table \ref{agiqa-3k} presents the performance results of different IQA models on AGIQA-3K database, demonstrating that \name\ shows strong performance. As we can see, \name\ achieves PLCC, SRCC, KRCC values of 0.8978, 0.8618 and 0.6776, respectively. These results outperform all compared methods, showcasing the great potential of our approach.

\begin{table}
\centering
\caption{Comparison with the state-of-the-art IQA methods on AGIQA-3K dataset. The best performance results are marked in {\color{red}RED} and the second-best performance results are marked in {\color{blue}BLUE}}
\renewcommand{\arraystretch}{1.3}
\setlength\tabcolsep{4pt}
\begin{tabular}{@{}c|cccc@{}}
\toprule
\textbf{Methods} & \textbf{PLCC} & \textbf{SRCC} & \textbf{KRCC} \\ \midrule
FID~\cite{heusel2017gans}     & 0.1860           & 0.1733      & 0.1158 \\
CEIQ~\cite{yan2019no}     & 0.4166           & 0.3228      & 0.2220 \\
NIQE~\cite{mittal2012making}     & 0.5171           & 0.5623      & 0.3876 \\
GMLF~\cite{xue2014blind}       & 0.8181           & 0.6987      & 0.5119 \\
CNNIQA~\cite{kang2014convolutional}      & 0.8469           & 0.7478      & 0.5580 \\
DBCNN~\cite{zhang2018blind}      & {\color{blue} 0.8759}    & 0.8207      & 0.6336 \\ 
CLIP-IQA~\cite{wang2023exploring}      & 0.8053           & {\color{blue} 0.8426}      & {\color{blue} 0.6468} \\ 
\textbf{CLIPAGIQA(Ours)}         & {\color{red} 0.8978}      & {\color{red} 0.8618}      & {\color{red} 0.6776}
\\\bottomrule
\end{tabular}
\label{agiqa-3k}
\end{table}

Table \ref{aigciqa2023} shows the comparison between our \name\ and other IQA methods on the AIGCIQA2023 dataset. It can be seen that our method not only meets or exceeds state-of-the-art performance in evaluating the quality of generated images but also significantly outperforms other IQA models in assessing the authenticity of the dataset, which refers to the ability to evaluate whether an image is AI-generated. This indicates that our model excels not only in quality assessment but also has great potential to extend to other aspects of evaluating generated images.

\begin{table}
\centering
\caption{Comparison with the state-of-the-art IQA methods on AIGCIQA2023 dataset. The best performance results are marked in {\color{red}RED} and the second-best performance results are marked in {\color{blue}BLUE}}
\renewcommand{\arraystretch}{1.3}
\setlength\tabcolsep{4pt}
\begin{tabular}{c|ccc|ccc}
\toprule
 & \multicolumn{3}{|c|}{\textbf{Quality}} & \multicolumn{3}{|c}{\textbf{Authenticity}} \\ \midrule
\textbf{Methods} & \textbf{PLCC} & \textbf{SRCC} & \textbf{KRCC}  & \textbf{PLCC} & \textbf{SRCC} & \textbf{KRCC} \\ \midrule
NIQE~\cite{mittal2012making}     & 0.5218 & 0.5060 & 0.3420 & 0.3954 & 0.3715 & 0.2453 \\
BRISQUE~\cite{mittal2012no}     & 0.6389 & 0.6239 & 0.4291 & 0.4796 & 0.4705 & 0.3142 \\
HOSA~\cite{xu2016blind}     & 0.6561 & 0.6317 & 0.4311 & 0.4985 & 0.4716 & 0.3101 \\
CNNIQA~\cite{kang2014convolutional}       & 0.7937 & 0.7160 & 0.4955 & 0.5734 & 0.5958 & 0.4085 \\
Resnet18~\cite{he2016deep}      & 0.7763 & 0.7583 & 0.5360 & 0.6528 & 0.6701 & 0.4740 \\
VGG16~\cite{simonyan2014very}      & 0.7973    & {\color{blue} 0.7961}      & {\color{blue} 0.5843 } & {\color{blue} 0.6807} & 0.6660 & 0.4813\\ 
VGG19~\cite{simonyan2014very}      & {\color{red} 0.8402} & 0.7733 & 0.5376 & 0.6565 & {\color{blue} 0.6674} & {\color{blue} 0.4843} \\ 
\textbf{CLIPAGIQA(Ours)}         & {\color{blue} 0.8302}      & {\color{red} 0.8140}      & {\color{red} 0.5991} & {\color{red} 0.7797} & {\color{red} 0.7940} & {\color{red} 0.5849}
\\\bottomrule
\end{tabular}
\label{aigciqa2023}
\end{table}

\begin{figure}
\includegraphics[width=\textwidth]{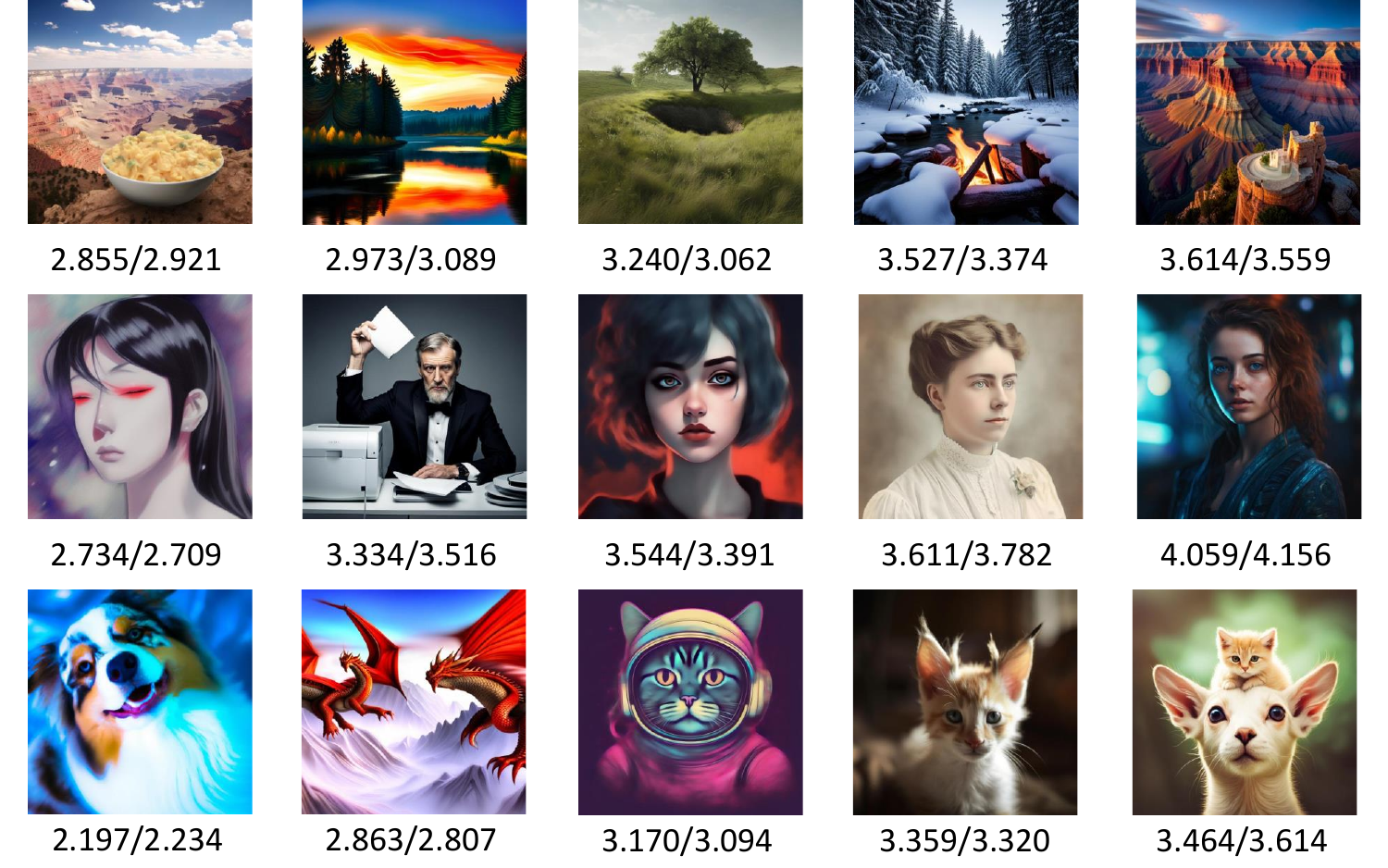}
\caption{\name\ for assessing overall perceptual quality. Left: Model Scores, Right: Human Scores} \label{fig3}
\end{figure}

Fig.\ref{fig3} shows that \name\ is able to assess overall perceptual quality to a level comparable to human judgment. It can assign reasonable scores based on the quality of the generated images. Notably, this model demonstrates several interesting capabilities. For instance, in the first column of the first row, where a strange bowl appears in the scenery image, it identifies common flaws in generated images and assigns a low score. Similarly, although the person in the second column of the second row looks lifelike, the model may detect subtle defects such as issues with the fingers and assigns a relatively low score. The first and second column of the third row also receive a low score maybe due to unrealistic elements and detail issues.

\subsection{Ablation Studies}\label{sub2}
As described in Section \ref{sub1}, we make three unique modifications to adapt CLIP for the quality assessment task. In this section, to verify the effectiveness of the proposed key components, we train five variants of \name\ in AGIQA-3K:

I) Without regression and using cosine similarity instead (following CoOp and using classification loss for tuning the context);
II) Changing the backbone network;
III) Changing the length of learnable contexts;
IV) Changing the length of quality categories;
V) Changing the type of quality categories.

\begin{table}[h]
\centering
\caption{Ablation Study Results}
\renewcommand{\arraystretch}{1.5}
\setlength\tabcolsep{4pt}
\begin{tabular}{@{}c|c|c|c|c|c@{}}
\hline
\textbf{No.} & \textbf{Ablation} & \textbf{Setting} & \textbf{PLCC} & \textbf{SRCC} & \textbf{KRCC} \\ \hline
0 & full model & ViT-B/16, 16, 6 adjectives & \textcolor{red}{0.8978} & \textcolor{red}{0.8618} & \textcolor{red}{0.6776} \\ \hline

1 & without regression & ViT-B/16, 16, 6 adjectives & 0.8183 & 0.8201 & 0.6693 \\ \hline
2 & - (backbone) & \textcolor{blue}{ViT-B/32}, 16, 6 adjectives & 0.8954 & 0.8614 & 0.6751 \\ 
&  & \textcolor{blue}{ResNet-101}, 16, 6 adjectives & 0.8837 & 0.8544 & 0.6665 \\ 
3 & - (context length) & ViT-B/16, \textcolor{blue}{8}, adjective & 0.8951 & 0.8595 & 0.6746 \\ 
  &  & ViT-B/16, \textcolor{blue}{32}, 6 adjectives & 0.8962 & 0.8605 & 0.6751 \\ \hline
4 & - (category length) & ViT-B/16, 16, \textcolor{blue}{8} adjectives & 0.8962 & 0.8616 & 0.6766 \\ 
5 & - (category type) & ViT-B/16, 16, 6 \textcolor{blue}{scores} & 0.8958 & 0.8604 & 0.6747 \\ \hline
\end{tabular}
\label{ablation}
\end{table}

The results indicate that removing or changing any single factor leads to a decrease in performance, confirming their contribution to the performance results in Table \ref{ablation}. It is worth noting that CLIP-IQA$^{\small +}$~\cite{wang2023exploring} has already validated the importance of learnable context and quality categories, so we only test the impact of regression on CLIP in the quality assessment of generated images. In variant 1, we observed a significant improvement when regression is added. This indicates that the combination of CLIP priors with a simple regression model is already effective.\par
In variants 2-5, although the impact on the model's performance is minimal, exploring these variants still provides us with valuable insights to understand and improve \name. Variants 2 and 3 are set up similarly to those explored in CoOp~\cite{zhou2022learning}. In our investigation of the backbone, we find a similar conclusion: the more advanced the backbone, the better the performance. However, the conclusion from CoOp that having more context tokens leads to better performance is not satisfied when the context length increased from 16 to 32. This can be due to the increased number of parameters making it harder for the model to converge to an appropriate state, warranting further investigation in future work. Additionally, we demonstrate that a “good” initialization does not make much difference, though this is not explicitly included in the table.\par
In variants 4 and 5, when the length of quality categories increases indefinitely, the task intuitively becomes a one-to-one classification task, yet the performance does not improve. Possible reasons could be that having too many quality categories makes synonyms indistinguishable, or the model parameters are insufficient to differentiate between categories. Changing the type of quality categories to numbers representing score relationships results in a performance drop, likely because CLIP rarely uses numbers in training, making it difficult to directly represent score magnitudes with numbers.

\section{Conclusion}
In this paper, we propose \name, a model that effectively adapts to new assessment requirements for generated images by leveraging CLIP's comprehensive visual and textual knowledge. Directly using CLIP has limitations and does not align well with the task of generated image quality assessment. To address this, we design various categories representing different quality levels to input into CLIP's text encoder, mitigating semantic ambiguities. By introducing a learnable prompts strategy and utilizing multiple quality-related auxiliary categories, we fully exploit CLIP's textual knowledge. Our regression network directly maps CLIP features to quality scores, effectively combining CLIP's capabilities with the task of generated image quality assessment, thereby enhancing the model's performance. Experiments demonstrate that \name, when trained with different datasets, performs excellently in both datasets. Ablation studies confirm the effectiveness of the proposed components. In the future, we will further improve our work by developing CLIP's own weights during training or by using multiple learnable contexts to explore multi-dimensional, fine-grained quality scores.

\section{Acknowledgments}
This work is supported by the National Natural Science Foundation of China (NSFC) under Grant No. 62272460, Beijing Natural Science Foundation under Grant No. 4232037.

\bibliographystyle{splncs04}
\bibliography{mybibliography}

\end{document}